# Proposal and study of statistical features for string similarity computation and classification

E.O. Rodrigues*, D. Casanova, M. Teixeira,
V. Pegorini and F. Favarim

Academic Department of Informatics,
Universidade Tecnológica Federal do Paraná (UTFPR),
Paraná, Brazil
Email: erickrodrigues@utfpr.edu.br
Email: dalcimar@utfpr.edu.br
Email: marceloteixeira@utfpr.edu.br
Email: vinicius@utfpr.edu.br
Email: favarim@utfpr.edu.br
*Corresponding author

E. Clua and A. Conci

Department of Computer Science,
Universidade Federal Fluminense (UFF),
Rio de Janeiro, Brazil
Email: esteban@ic.uff.br
Email: aconci@ic.uff.br

Panos Liatsis

Department of Electrical Engineering and Computer Science,
Khalifa University,
Abu Dhabi, UAE
Email: pliatsis@pi.ac.ae

**Abstract:** Adaptations of features commonly applied in the field of visual computing, co-occurrence matrix (COM) and run-length matrix (RLM), are proposed for the similarity computation of strings in general (words, phrases, codes and texts). The proposed features are not sensitive to language related information. These are purely statistical and can be used in any context with any language or grammatical structure. Other statistical measures that are commonly employed in the field such as longest common subsequence, maximal consecutive longest common subsequence, mutual information and edit distances are evaluated and compared. In the first synthetic set of experiments, the COM and RLM features outperform the remaining state-of-the-art statistical features. In 3 out of 4 cases, the RLM and COM features were statistically more significant than the second best group based on distances (P-value < 0.001). When it comes to a real text plagiarism dataset, the RLM features obtained the best results.

**Biographical notes:** E.O. Rodrigues is a Professor of Computer Science at the Universidade Tecnologica Federal do Parana, Brazil. He received several awards during his graduation which includes two awards for the best PhD thesis in exact sciences, an honourable mention from CAPES for his PhD thesis, being one of the top three best PhD thesis in computer science in Brazil and a best MSc dissertation in exact sciences. Previously, he worked as a visiting graduate research assistant at Khalifa University, UAE. He has published papers in top tier journals and has been working with computer vision, data mining, optimisation and biological/medical informatics.

D. Casanova has received his BS in Computer Science from the University of Western Santa Catarina UNOESC (2005), Master's in Computer Science and Computational Mathematics by Institute of Mathematics and Computer Sciences ICMC-USP (2008), PhD in Computational Physics by Institute of Physics of São Carlos IFSC-USP (2013) and post-PhD at the Institute of Physics of São Carlos IFSC-USP. He is currently a Professor at the Federal University of Technology – Paraná UTFPR. He has experience in computer science, computational physics, and applications multidisciplinary areas, mainly in the following topics: computer vision, complex networks, machine learning, deep learning, and bioinformatics.

M. Teixeira received his BSc in Computer Science, MSc in Computer Engineering, and PhD in Automation and Systems Engineering. Since then, he has been a faculty member of the Department of Informatics at the Federal University of Technology Paraná, Brazil, where he is also a member of the Graduate Program in Electrical Engineering. His current research interests include discrete event systems, manufacturing systems, cyber physical systems, and synthesis of controllers for industrial automation.

V. Pegorini received his Master's in Electrical Engineering from the Universidade Tecnológica Federal do Paraná, Pato Branco, PR, Brazil in 2015. He is currently an Assistant Professor in the Department of Informatics at the Universidade Tecnológica Federal do Paraná, Pato Branco, PR, Brazil. His current research interests are machine learning and computer vision.

F. Favarim received his BSc in Computer Science (2000), Master's (2003) and PhD (2009) in Electrical Engineering (Automation and Systems). In 2006, he participated in an internship in LASIGE research unit/navigators research team in the Department of Informatics of the University of Lisboa Faculty of Sciences, Portugal. Since 2009, he has been a faculty member of the Department of Informatics at the Federal University of Technology Paraná, Brazil. His current research interests include parallel and distributed systems, computer networks and internet of things.

E. Clua is a Professor of Universidade Federal Fluminense and coordinator of UFF Medialab. He was nominated young scientist of the State of Rio in 2009 and 2013, and in 2015 received the nomination of CUDA Fellow. His main research and development areas are digital games, virtual reality, GPUs, and visualisation. He is currently a coordinator of the NVIDIA Center of Excellence, which is located at the CS Institute of UFF.

A. Conci is a Civil Engineer with MSc and PhD in the same area, and Full Professor at Universidade Federal Fluminense (UFF). She works with computer modelling, computer vision, image analysis and bioinformatics. She oriented over 125 students and is member of the ISGG, ABCM, SBrT and SBC. She acts in the editorial office of a couple of international journals and has cooperated in terms of research with many scholars in various countries. She has a substantial amount of high quality publications (more than 3k citations, h-index = 29 and i10 = 75). She has funding from the Brazilian government for coordinating over 20 research projects.

Panos Liatsis is a Professor in the Department of Electrical Engineering and Computer Science at Khalifa University of Science and Technology. He received his Diploma in Electrical Engineering from the University of Thrace, Greece, and PhD in Electrical Engineering and Electronics from the University of Manchester, UK. He commenced his academic career at the University of Manchester, prior to joining City, University of London, UK, where he was a Professor and the Head of the Electrical and Electronic Engineering Department. His research interests are image processing, computer vision, pattern recognition and machine learning.

---

## 1 Introduction

Text analysis can be structured in a few different ways. In opinion and sentiment analysis (Godbole et al., 2007; Carvalho et al., 2014) texts are analysed so that a prevalent emotion is usually inferred and associated to it, e.g., anger, happiness, relief, and others. Some words can be compared to different templates or even be turned into features to be classified, which would predict the emotion.

Other applications very frequently include some degree of string similarity computation. In optical character recognition (OCR), scanned words are compared to a dictionary of predefined words, where the most similar word in the dictionary is selected. In text plagiarism, the degree of similarity among texts is analysed and used to infer distinct degrees of plagiarism. Even in text entailment, which evaluates the relation of sections of a text, fragments of the text can be compared to possible subsequent text fragments using similarity measures.

In this work, adaptations of the

1. co-occurrence matrix (COM)
2. run-length matrix (RLM)
3. weighted mutual information (WMI), all of which can be found in more detail in Rodrigues et al. (2016), are proposed and used in string similarity computation.

Features that stem from these structures are often employed in the field of visual computing but were never used to compute string similarity before. Extracted features are analysed using classification algorithms (Rodrigues et al., 2018), which subsequently determine the degree of similarity between two strings (words, fragments of texts, etc.) that are being compared. Figure 1 highlights the proposed framework.

**Figure 1** Proposed framework for string comparison (see online version for colours)

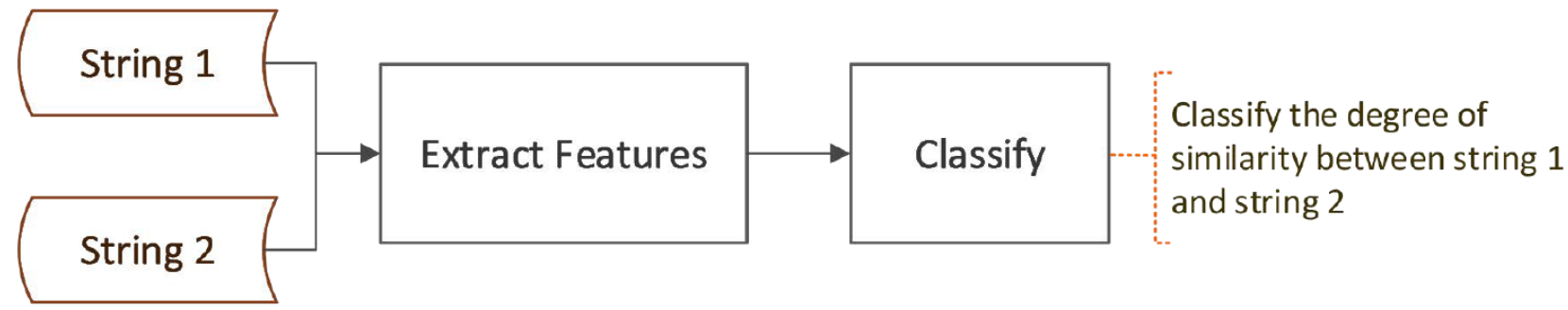

## 1.1 Notation

Several string operators are utilised in this work such that a formal definition for each employed operator is required. Strings are defined as a concatenation of characters $c_0, \ldots, c_{n-1}$, where $n$ represents their length. A string $w$ is defined as:

$$w = \{c_0, \ldots, c_{|s|-1}\} \tag{1}$$

where the |.| operator also represents the length of the respective string. The character of string w at position $p$ is given by $w[p]$, where $0 \leq p < |w|$.

A substring of $w$ is obtained using $w[k_i, k_f)$, where [ represents inclusion and ) represents exclusion. For a given string $w = c_0c_1c_2c_3$, its substring $w[1, 3)$ is given by $w[1, 3) = c_1c_2$, as shown:

$$\begin{aligned} &w = c_0c_1c_2c_3 \\ &w[1,3) = c_1c_2 \\ &w[2] = c_2 \end{aligned} \tag{2}$$

Similarly, function $atr(w, p)$ returns and removes from $w$ the character at position $p$. In contrast to $w[p]$, $atr$ removes the character from $w$. Let us assume that $w = c_0c_1$, after calling $atr(w, 1)$, $w$ is equal to $c_0$.

The operator intersection $\cap$ returns the intersection of characters between two strings $w_1$ and $w_2$ (the order does not matter):

$$\begin{aligned} &w_1 = c_0c_1c_3 \\ &w_2 = c_3c_1 \\ &w_1 \cap w_2 = c_1c_2 \end{aligned} \tag{3}$$

Two strings $w_1$ and $w_2$ are concatenated using the symbol $\oplus$, where $w_2$ is appended at the end of $w_1$:

$$\begin{aligned} &w_1 = c_0c_1c_3 \\ &w_2 = c_3c_1 \\ &w_1 \oplus w_2 = c_0c_1c_3c_3c_1 \end{aligned} \tag{4}$$

The operator $rem(w)$ removes the last character of string $w$, such that

$$
\begin{aligned}
&w = c_0c_1c_2c_3 \\
&rem(w) = c_0c_1c_2
\end{aligned} \tag{5}
$$

Other functions are also employed, $qnt(s \in w)$ returns the occurrences of string $s$ in string $w$, where the order is respected:

$$
\begin{aligned}
&w = c_0c_1c_2c_3c_4c_0c_1 \\
&s = c_0c_1 \\
&qnt(s \in w) = 2
\end{aligned} \tag{6}
$$

Function $rep(w, p, c)$ replaces the character of string $w$ at position $p$ with character $c$:

$$
\begin{aligned}
&w = c_0c_1c_2c_3c_4c_0c_1 \\
&rep(w, 1, c_2) = c_0c_2c_2c_3c_4c_0c_1
\end{aligned} \tag{7}
$$

which is usually used throughout the work in conjunction with function $ran(k_i, k_f)$ that returns a random number between $k_i$ and $k_f$, both inclusive.

Finally, function $add(w, c)$ takes the character $c$ and appends it to the beginning (position 0) of string $w$. Similarly, $addr(w, c)$ takes $c$ and appends it either to the beginning or ending of string $w$, with equal probability.

$$
\begin{aligned}
&w = c_0c_1c_2c_3c_4 \\
&add(w, c_1) = c_1c_0c_1c_2c_3c_4 \\
&addr(w, c_1) = \begin{cases} c_1c_0c_1c_2c_3c_4, & \text{50\% probability} \\ c_0c_1c_2c_3c_4c_1, & \text{otherwise} \end{cases}
\end{aligned} \tag{8}
$$

## 2 Literature review

Islam and Inkpen (2008) aimed to evaluate the similarity of texts employing the normalised longest common subsequence (NLCS) and maximal consecutive longest common subsequence (MCLCS). NLCS and MCLCS are given by equations (9) and (10), respectively,

$$
NLCS(w_1, w_2) = \frac{|LCS(w_1, w_2)|^2}{|w_1||w_2|} \tag{9}
$$

$$
NMCLCS_{n_1,n_2}(w_1, w_2) = \frac{|MCLCS_{n_1,n_2}(w_1, w_2)|^2}{|w_1||w_2|} \tag{10}
$$

where LCS stands for the longest common subsequence. An arbitrary string $w_3$ is a longest common subsequence of strings $w_1$ and $w_2$ if $w_3$ is a common subsequence of $w_1$ and $w_2$ of maximal length, i.e., there is no common subsequence of $w_1$ and $w_2$ that has greater length (Hirchberg, 1977).

Variables $n_1$ and $n_2$ in equations (10) and (11) stand for the starting position of the MCLCS in $w_1$ and $w_2$, respectively. For instance, if $n_1 = 0$, the sequences at the first character of $w_1$ are considered. Therefore, if $n_1 = 2$, the first two characters of $w_1$ are erased or ignored and we start at the third. The same holds for $n_2$. If $n_2 = 1$, the subsequence search starts with the character at position 1 of $w_2$. In other words, the first character of $w_2$ is erased or ignored.

The MCLCS ($MCLCS_{n_1,n_2}$) is given by equation (11), where $rem(w_1)$ removes the last element of the string $w_1$, in this specific case, if and only if $|w_1| > |w_2|$. In the case of $rem(w_2)$, the last element of $w_2$ is removed if $|w_2| > |w_1|$. Given the instances $w_1 = olvahirah$ and $w_2 = oliveira$, then $LCS(w_1, w_2) = olvira$, $MCLCS_{0,0}(w_1, w_2) = ol$, and $MCLCS_{\forall n_1, \forall n_2} = ira$.

$$MCLCS_{n_1,n_2}(w_1, w_2) = \begin{cases} w_1, & \text{if } w_1 = w_2 \\ MCLCS_{n_1,n_2}(rem(w_1), rem(w_2)), & \text{otherwise} \end{cases} \tag{11}$$

Mutual information (Turney, 2002) is shown in equation (12), where $g$ is usually 2, 10 or $e$. This equation illustrates MI in the context of this work, where $c_1$ and $c_2$ are characters of strings $w_1$ and $w_2$, respectively.

$$MI(w_1, w_2) = \sum_{c_1 \in w_1} \sum_{c_2 \in w_2} \rho(c_1, c_2) \log_g \frac{\rho(c_1, c_2)}{\rho(c_1)\rho(c_2)} \tag{12}$$

The value $\rho(c_1, c_2)$ represents the joint probability of the element $c_1$ co-occurring with element $c_2$. In this context, a co-occurrence is defined as an occurrence of the same character in the same position in both strings (e.g., $w_1[p] = w_2[p]$, at all possible positions $p$).

Given strings $w_1 = c_0c_1c_1c_3$ and $w_2 = c_0c_1c_1c_4$, the probability of occurrence of $c_1$ and $c_1$ is $\rho(c_1, c_1) = 2/4$. That is, two co-occurrences of $c_1$ with $c_1$ in a total of four co-occurrences. Similarly, $\rho(c_3, c_3) = 0$, they never co-occur. $\rho(c_1)$ and $\rho(c_2)$ stand for the marginal probability of $c_1$ occurring in $w_1$ and $c_2$ occurring in $w_2$, respectively.

Apart from LCS, MCLCS and mutual information, edit distances (Rodrigues, 2018) are also used to compare strings with no language bias. Hamming, Levenshtein and Damerau distances were evaluated in this work. Although the dice coefficient cannot be defined as a distance, we have also considered it and included this one in the distances group for ease of analysis.

The Hamming distance (Bookstein et al., 2002) employed in this work is a slight modification of the original one. At first, we transform the longest string in order to match its length to the other. Exceeding characters of the longest string are disregarded. Given $w_1 = example$ and $w_2 = ejamjletwo$, $w_2$ is transformed to $w_2 = ejamjle$. Just after that, the function *ModHam* in equation (13) is computed:

$$ModHam(w_1, w_2) = |w_1| - |w_1 \underset{\times}{\cap} w_2| \tag{13}$$

in this specific case, $\underset{\times}{\cap}$ represents the number of characters that are the same in both strings, differing from the previously defined intersection operation $\cap$.

As opposed to Hamming, which considers substitutions, Levenshtein distance (Schulz and Mihov, 2002) also considers insertions and deletions on the edit distance computation. The Levenshtein distance $Lev_{w_1,w_2}(len(w_1), len(w_2))$ is given by equation (14),

$$Lev_{w_1,w_2}(i,\ j) = \begin{cases} \max(i,\ j), & \text{if } \min(i,\ j) = 0 \\ MIN_M, & \text{otherwise} \end{cases} \tag{14}$$

respecting,

$$MIN_M = \min \begin{cases} Lev_{w_1,w_2}(i-1,\ j)+1 \\ Lev_{w_1,w_2}(i,\ j-1)+1 \\ Lev_{w_1,w_2}(i-1,\ j-1)+q \end{cases} \tag{15}$$

where $q$ assumes 0 if the character at position $i$ in $w_1$ is equal to the character at position $j$ in $w_2$ and 1 otherwise (i.e., 1 if $w1[i] \neq w_2[j]$ and 0 otherwise).

Damerau improves the Leveshtein distance by adding transposition or swap operations. Therefore, Damerau distance (Wieling et al., 2009) ($Dam_{w_1,w_2}(len(w_1), len(w_2))$, also known as Damerau-Levenshtein distance) includes substitutions, insertions, deletions and swaps of two adjacent characters as shown in equation (16).

$$Dam_{w_1,w_2}(i,\ j) = \begin{cases} 0 & \text{if } i = j = 0 \\ Dam_{w_1,w_2}(i-1,\ j)+1 & \text{if } i > 0 \\ Dam_{w_1,w_2}(i,\ j-1)+1 & \text{if } j > 0 \\ Dam_{w_1,w_2}(i-1,\ j-1)+q & \text{if } i,\ j > 0 \\ Dam_{w_1,w_2}(i-2,\ j-2)+1 & \begin{Bmatrix} \text{if } i, i > 1 \text{ and} \\ w_1[i] = w_2[j-1] \text{ and} \\ w_1[i-1] = w_2[j] \end{Bmatrix} \end{cases} \tag{16}$$

Similarly to the previous equation, $q$ assumes 0 if the character at position $i$ in $w_1$ is equal to the character at position $j$ in $w_2$ and 1 otherwise.

At last, the dice coefficient is defined in equation (17).

$$Dice(w_1,\ w_2) = \frac{2|w_1 \cap w_2|}{|w_1| + |w_2|} \tag{17}$$

These described features are the popular statistical features when it comes to string analysis and string comparison. However, several works consider semantical information for comparing strings. The work of Han et al. (2013) use polysemy to improve the MI measure. Dagan et al. (19995) use a similarity network along with heuristics to link words to a degree of similarity. Corley and Mihalcea (2005) employ several measures that consider semantic information including taxonomy and dictionaries. Lin and Sandkuhl (2008) also address some taxonomy on the same context.

Although these measurements in fact improve the efficiency of the comparison, they are very frequently tied and biased towards a single language. Different languages have different words, semantics and structures. Therefore, for every single language a new taxonomy or dictionary has to be created. In this work, just statistical and/or rule-base measures are considered, enabling the approach to work on any language.

# 3 Proposed methodology

The usual mutual information measure was slightly altered to be sensitive to the ordering of the characters. We call this modification WMI. A similar version has been already used in visual recognition tasks, outperforming plain mutual information such as in a previous work (Rodrigues et al., 2016). The adapted pointwise weighted mutual information (PWMI) between character $c_1$ of string $w_1$ and $c_2$ of string $w_2$ is shown in equation (18), where m is a constant > 1 (it could be < 1 if measuring dissimilarity) and adds the weight to the equation (*m* was set to 2 in the experiments), and *d* is a distance by which the longest string is shifted. For instance, if $d = 1$ and the longest string is *abc*, then it would be shifted to *bca* before PWMI is computed. The initial *d* characters are moved to the end of the string.

$$PWMI_d\left(c_1, c_2\right) = \begin{cases} m \log_g \dfrac{\rho\left(c_1, c_2\right)}{\rho\left(e_1\right)\rho\left(c_2\right)}, & \text{if } c_1 = c_2 \\ \log_g \dfrac{\rho\left(c_1, c_2\right)}{\rho\left(c_1\right)\rho\left(c_2\right)}, & \text{otherwise} \end{cases} \tag{18}$$

The shift operation is shown in equation (19). Besides being used in WMI, this SHIFT operation was also used in the common MI measure.

$$SHIFT(w, d) = w\left[d, |w|\right) \oplus w\left[0, d\right) \tag{19}$$

In order to generate a total measurement of all the possible displacement distances, it is just necessary to perform the summations as shown by the proposed PWMIS in equation (20).

$$PWMIS\left(w_1', w_2'\right) = \sum_{d=0}^{\max\left(|w_1'|-1, |w_2'|-1\right)} PWMI_d\left(c_1, c_2\right) \forall c_1 \in w_1', c_2 \in w_2' \\ \Big| w_1' = SHIFT\left(w_1', d\right), w_2' = SHIFT\left(w_2', d\right) \tag{20}$$

The idea behind the adaptation is that common mutual information does not account or differentiate exactly equal strings, it just considers the probability of co-occurrences, disregarding the characters themselves. For instance, $S_1$ = '213', $S_2$ = '321' and $S_3$ = '321' would yield the same mutual information: $MI(S_1, S_2) = MI(S_2, S_3)$, while $PWMIS(S_1, S_2) < PWMIS(S_2, S_3)$ as long as $m > 1$.

## *3.1 Features derived from COMs*

In visual computing, COMs (Rodrigues et al., 2014) contain the co-occurrences of pixel grey values at a given distance. Figure 2 illustrates a COM computed from an input image. The values representing the rows and columns are the possible grey values of the pixels. In this case, the chosen distance is (1, 0), which is analogous to the pixel at the right. The position (1, 2) of the matrix has the value 4 due to four co-occurrences respecting the distance (1, 0) of the values {1, 2} in the image (highlighted in red).

**Figure 2** Construction of a COM (see online version for colours)

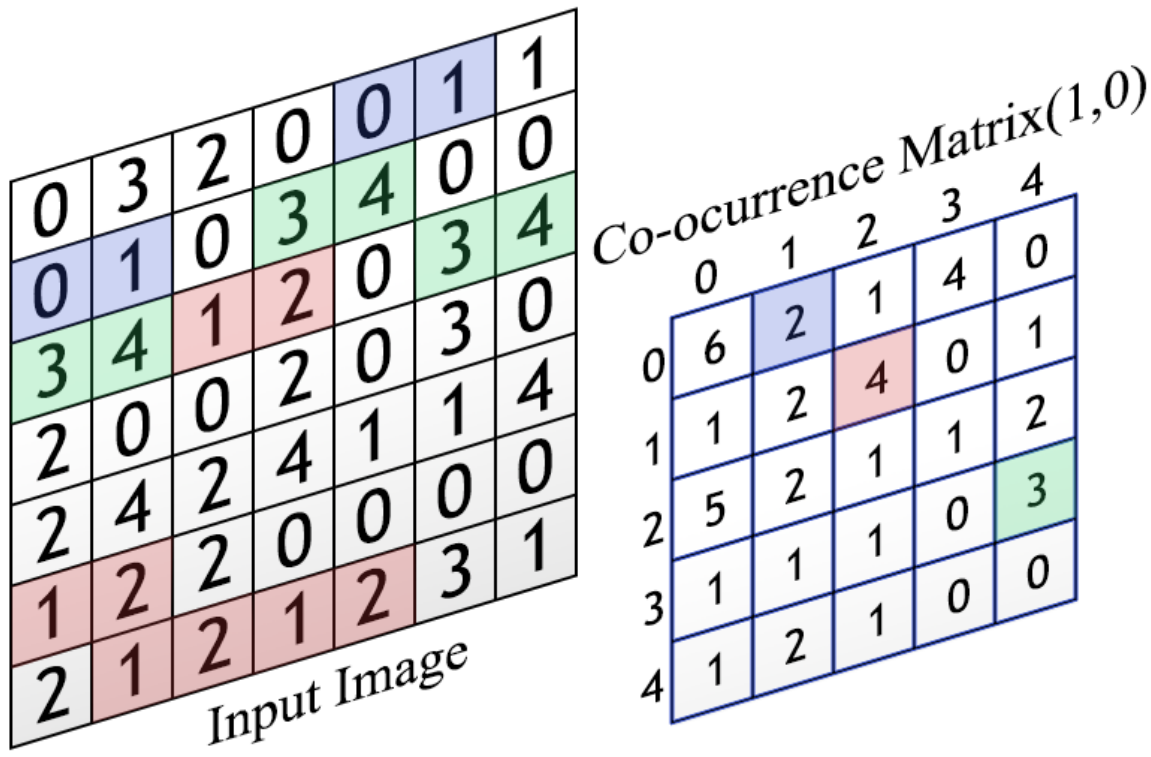

This COM adaptation considers a distance parameter (in this case the orientation can only be from left to right), and considers just co-occurrences of equal elements. For instance, the strings *aaabb* and *aaabc* contain three co-occurrences of *a* and one co-occurrence of *b* at distance 0. The counting is performed always from the first string $w_1$ to the second one $w_2$. The formulation is given by equation (21). $COM(w_1, w_2, p, d)$ returns the number of co-occurrences of the character at position $p$ in $w_1$ between words $w_1$ and $w_2$, given a certain distance $d$.

$$COM\left(w_1, w_2, p, d\right) = \sum_{k=0}^{|w_1|-1} \begin{cases} 1, & \text{if } w_1[p] = w_1[k] = w_2[k+d] \\ 0, & \text{otherwise} \end{cases} \tag{21}$$

Figure 3 illustrates the example given in the previous paragraph. This case considers $p = 0$ (or $a$) for the value 3 and $p = 1$ (or $b$) for value 1 in the matrix, and both consider $d = 0$. The character '$a$' in $w_1$ also occurs three times in $w_2$ when the distance 0 is fixed. The character '$b$' occurs once. We are not interested in counting characters that co-occur but are not the same, despite the fact that this idea is in the original COM (for images). The cell where $b$ occurs with $c$ in Figure 3 would be marked with 1 if we were to use the original co-occurrence idea. Although we do not employ this idea in this work, this type of co-occurrence could be further investigated while considering characters (or even words) that inherit degrees of relation among them, which could be interesting and improve even more the results. However, for this work, this type of co-occurrence is ignored as we are pursuing pure statistical methods and not techniques based on language.

**Figure 3** Construction of a COM for strings (see online version for colours)

input

$w_1$
*aaabb*

$w_2$
*aaabc*

co-occurrence

| | a | b | c |
|---|---|---|---|
| a | 3 | 0 | 0 |
| b | 0 | 1 | 0 |
| c | 0 | 0 | 0 |

Some features can be extracted from equation (21). As an example, the co-occurrence probability (COP) of the character at position $p$ in $w_1$, regarding words $w_1$ and $w_2$ and distance $d$, is given by equation (22). ACOP for all distances can be computed using equation (23).

$$COP(w_1, w_2, p, d) = \frac{COM(w_1, w_2, p, d)}{|w_1|} \tag{22}$$

$$COP(w_1, w_2, p) = \frac{\sum_{d=0}^{|w_1|-1} COM(w_1, w_2, p, d)}{|w_1|} \tag{23}$$

A feature to address COP as a whole was also created. This feature, called probability score (PS), computes a score for each element in $w_2$. If the element is equal to any element of $w_1$ at distance $d$, then its COP is summed to the score. Otherwise, a penalty equal to –1 is added to it. The formulation is shown in equation (24).

$$PS(w_1, w_2, d) = \sum_{p=0}^{|w_2|-1} \begin{cases} COP(w_1, w_2, p_{w_1}, d), & \text{if } w_2[p] \in w_1 \\ -1, & \text{otherwise} \end{cases} \tag{24}$$

where $p_{w_1}$ corresponds to the position in $w_1$ of the character at position $p$ in $w_2$, i.e., $w_1[p_{w_1}] = w_2[p]$.

Equation (24) returns a score for a specific distance. The total probability score (TPS), in equation (25), computes PS for all possible distances.

$$TPS(w_1, w_2) = \sum_{d=0}^{|w_2|-1} \sum_{p=0}^{|w_2|-1} \begin{cases} COP(w_1, w_2, p_{w_1}, d), & \text{if } w_2[p] \in w_1 \\ -1, & \text{otherwise} \end{cases} \tag{25}$$

Finally, a measurement that considers the equilibrium of the co-occurrences over distances (CODs) is shown in equation (26). This feature is called CODs, where g can assume any value (in the experiments it was set to 1). In some cases it would be interesting to consider $g = d$ or $g = 1/d$ in order to weight the distances appropriately. That is, if it is desired to compare similar words, then smaller distances $d$ should be favoured by using $g = 1/d$, for instance. Otherwise, $g = d$ would be the suitable choice.

$$COD(w_1, w_2) = \sum_{d=0}^{\lfloor |w_1|/2 \rfloor - 1} \left( \sum_{p=0}^{|w_1|-1} COM(w_1, w_2, p, d) - \sum_{p=0}^{|w_1|-1} COM\left(w_1, w_2, p, d + \lceil |w_1|/2 \rceil\right) \right)^g \tag{26}$$

## 3.2 Features derived from RLMs

Beyond these previously presented features based on the COM, an adaptation of another matrix that is commonly used in visual computing is proposed, which is the RLM (Tang, 1998; Rodrigues et al., 2016). RLM is based on the run length encoding, which is commonly used to compress texts. For instance, a sequence of characters like: *aaabbccc* is written as *3a2b3c* if the run-length encoding is applied. The three occurrences of *a* is substituted with a number 3 along with the character that is being repeated.

In visual computing, each column of the RLM represents the length of the run, and each line represents the grey level of the pixel. Figure 4 illustrates a RLM computed from an input image at orientation 0°, which corresponds to the horizontal orientation.

**Figure 4** Construction of a RLM (see online version for colours)

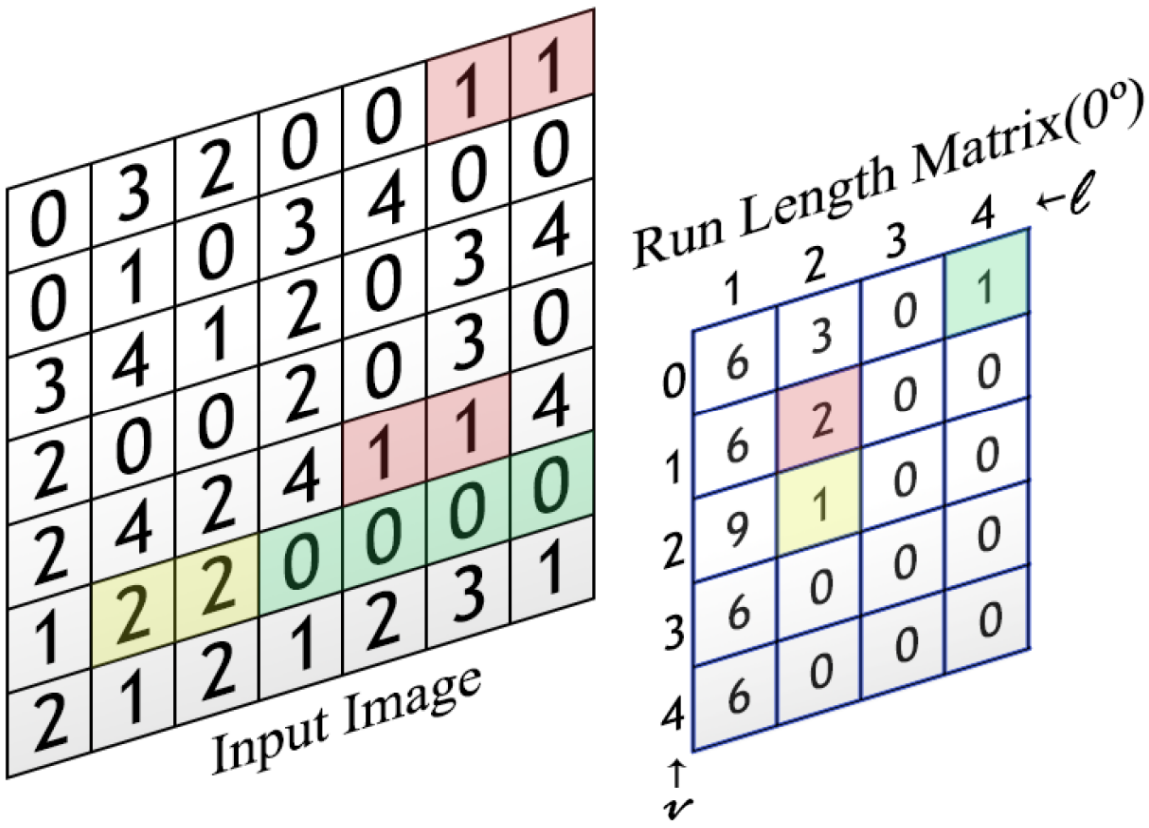

Similarly to the COM, the values in the RLM represent the amount of occurrences. For instance, the value ($v$, l) in the RLM matrix is equal to 6 if $v = 4$ and $l = 1$. This 6 represents the six appearances of the grey value 4 in the image with run length equal to 1, respecting the horizontal direction (0°). That is, the number 4 appeared six times alone in the image, respecting the given orientation. The adapted version of the RLM matrix for string comparison is shown in equation (27), where $qnt(s \in w_2)$ returns the amount of occurrences of the sequence s in the word $w_2$.

$$RLM(w_1, w_2, l) = qnt(s \in w_2)\ \forall s = w_1[p, p+l) \\ \big| 0 \le p < |w_1|, 0 < l < |w_1| - p \tag{27}$$

Figure 5 illustrates the run length vector for strings $w_1$ and $w_2$

**Figure 5** Construction of a RLM for strings

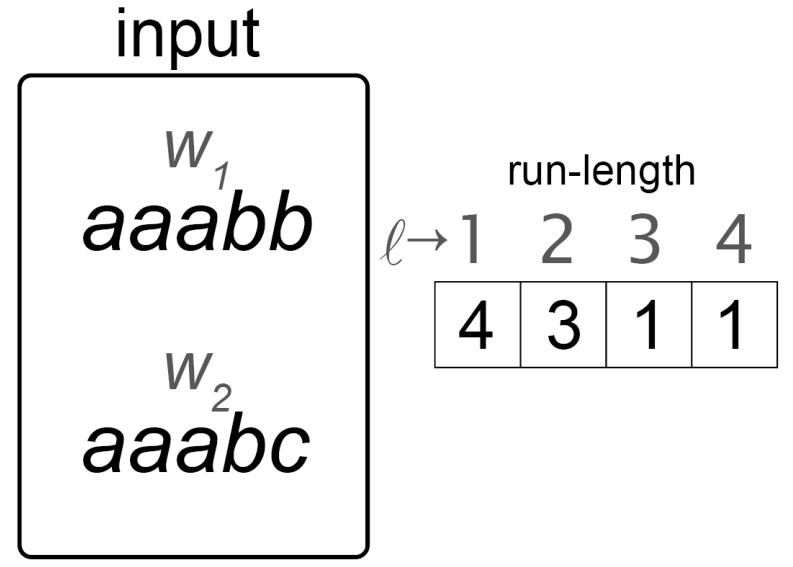

In Figure 5, for length 1, we have:

- Three occurrences of the sequence *a* in $w_2$.
- One occurrence of the sequence *b* in $w_2$.
- As *c* does not belong to $w_1$, it does not count.
- *Total:* $rlm(w_1, w_2, 1) = 3 + 1 = 4$.

For length 2, we have:

- Two occurrences of the sequence *aa* in $w_2$.
- One occurrence of the sequence *ab* in $w_2$.
- Zero occurrences of the sequence *bb* in $w_2$.
- *Total:* $rlm(w_1, w_2, 2) = 2 + 1 = 3$.

For length 3, we have:

- One occurrence of the sequence *aaa* in $w_2$.
- Zero occurrence of the sequence *abb* in $w_2$.
- *Total:* $rlm(w_1, w_2, 3) = 1$.

For length 4, we have:

- One occurrence of the sequence *aaab* in $w_2$.
- Zero occurrences of the sequence *aabb* in $w_2$.
- *Total:* $rlm(w_1, w_2, 4) = 1$.

Essentially, the proposed RLM computes the amount of sequences of $w_1$ that occurs in $w_2$ respecting a length l. In other words, given a certain length, we want to know the amount of occurrences in $w_2$ of sequences of this predefined length in $w_1$. As strings are one-dimensional, the produced RLM is a simple vector in this case.

One of the features that can be extracted from the RLM is the sum of occurrences (SO), given by equation (28).

$$SO(w_1, w_2) = \sum_{l=0}^{|w_1|-1} RLM(w_1, w_2, l) \tag{28}$$

A weight can also be applied to the SO equation, weighting positively words that have longer common subsequences, as shown in equation (29). The weight is given by variable $l$, and $g$ can assume any constant value. In this work, $g$ was equal to 1.

$$WSO(w_1, w_2) = \sum_{l=0}^{|w_1|-1} l^g \; RLM(w_1, w_2, l) \tag{29}$$

The maximal number of occurrences in the matrix was also selected as a feature, which is shown in equation (30) as maximal occurrence (MO), where $l_f$ is the maximal possible length of the matrix.

$$MO(w_1, w_2) = \max \begin{cases} RLM(w_1, w_2, 1) \\ \dots \\ RLM(w_1, w_2, l_f - 1) \\ RLM(w_1, w_2, l_f) \end{cases} \quad (30)$$

The length associated to the MO, called maximal occurred run length (MORL), is shown in equation (31).

$$MORL(w_1, w_2) = l \left| RLM(w_1, w_2, l) = MO(w_1, w_2), \quad 0 \le l \le |w_1| \right. \quad (31)$$

Finally, the MO equation can also be modified to return the maximal occurrences of minimal length (MOML) and the minimal length of maximal occurrences (MLMO) as shown in equations (32) and (33), respectively, assuming that $g$ is a constant big enough to properly weight the RLM values and small enough not to overflow variable data types.

$$MOML(w_1, w_2) = \max\left( \bigcup_{l=0}^{|w_1|-1} \frac{RLM(w_1, w_2, l)}{(l^g + 1)} \right)(l^g + 1) \quad (32)$$

$$MLMO(w_1, w_2) = \min(l)\ \forall l \left| 0 \le l < |w_1|, \quad \max\left(RLM(w_1, w_2, l)\right) \right. \quad (33)$$

The weight in equation (32) positively influences the choice of the minimal lengths by dividing the RLM values by a smaller number. The division effect is later removed by the multiplication of the same value after selecting the correct length ($\times (l^g + 1)$).

At last, a feature analogous to MCLCS can also be directly extracted from the RLM matrix, which is shown in equation (34).

$$RLMMCLCS(w_1, w_2) = \max\left( \bigcup_{l=0}^{|w_1|-1} \begin{cases} l, & \text{if } RLM(w_1, w_2, l) \ge 1 \\ 0, & \text{otherwise} \end{cases} \right) \quad (34)$$

## 4 Experimental results

Two major groups of experiments were performed. The first group consists of experimentations with a synthetic dataset, where two strings were generated according to a set of rules. The second group of experiments considers a real case plagiarism dataset, where the degree of copy of two texts is predicted. In both cases, the previously described features and classification algorithms are employed to perform the prediction.

### *4.1 Synthetic dataset experiments*

At first, a string $w_1$ of length up to 14 ($0 < |w_1| < 15$) was generated. Once this word is created, a second word is generated based on this former word. A sequence of random changes that may occur alter this first word. With probability $R$, the string is diminished according to equation (35). $R$ is a variable that can be increased to introduce more randomness in the generation, or decreased if the opposite effect is desired.

In what follows, several functions that alter the created string $w_1$ are presented. The order in which these functions should be applied to the string respects the order in which they are presented. $C_1(w)$ is the first function to be applied and is responsible for reducing the length of the initial string $w_1$.

$$C_1(w) = \begin{cases} w\left[0, ran\left(0, |w|\right)\right), & \text{if } ran(0,1) \geq 0.9 - R \\ w, & \text{otherwise} \end{cases} \tag{35}$$

After $C_1$, the returned string is then processed by $C_2$ for $|w| = 2$ times, $C_2$ is shown in equation (36).

$$C_2(w) = \begin{cases} rep\left(w, ran\left(0, |w|\right), ran\left(\text{‘}A\text{’}, \text{‘}z\text{’}\right)\right), & \text{if } ran(0,1) \geq 1 - R \\ w, & \text{otherwise} \end{cases} \tag{36}$$

The $rep(w, v_1, v_2)$ function in equation (36) replaces in string w the character at position $v_1$ with $v_2$, where ran(‘$A$’, ‘$z$’) returns a random character between $a$ and $z$, including lower and upper cases. In other words, function $C_2$ replaces a random character in the given string with a new randomly generated character.

The following modification function given by equation (37) shuffles the letters in the string, where function $atr(w, k)$ returns and removes the character of word $w$ at position $k$. Function $add(w, c)$ takes a character c and puts it in the first position of $w$ (position 0). Essentially, function $C_{3a}$ takes the first character of $w$ with 40% of probability or a character at a random position otherwise and adds it to the beginning of the word. This process is repeated until no elements are left in $w$. In this specific case, $R$ was disregarded.

$$C_{3a}(w) = \begin{cases} add\left(w, atr\left(w, ran\left(0, |w|-1\right)\right)\right), & \text{if } ran(0,1) \leq 0.6 \\ w, & \text{otherwise} \end{cases} \tag{37}$$

The function $C_{3a}$ is called only with 0.6 – $\sqrt{(R)}$ probability, as shown in the actual $C_3$ function in equation (38). If $ran(0, 1) < 0.6 - \sqrt{R}$ then the shuffle function $C_{3a}$ is not triggered.

$$C_3(w) = \begin{cases} C_{3a}(w), & \text{if } ran(0,1) \geq 0.6 - \sqrt{R} \\ w, & \text{otherwise} \end{cases} \tag{38}$$

The next modification function is shown in equation (39), which adds characters at the beginning or ending of $w$. For $|w| = 2$ times, $C_4$ in equation (40) is triggered. The function $addr(w, c)$ adds a character $c$ at the end or beginning of $w$ with equal probability.

$$C_{4a}(w) = \begin{cases} addr\left(w, ran\left(\text{‘}A\text{’}, \text{‘}z\text{’}\right)\right), & \text{if } ran(0,1) \geq 0.9 - R \\ w, & \text{otherwise} \end{cases} \tag{39}$$

$$C_4(w) = \begin{cases} C_{4a}(w), & \text{if } random(0,1) \geq 0.4 - R \\ w, & \text{otherwise} \end{cases} \tag{40}$$

**Table 1** Some randomly generated words after applying the CF function

| *Generated word (w1)* | *Changed word (w2)* |
|---|---|
| cakE_Cg_kakG[h | GcaKe_Co |
| JacKEts | pJacKIs |
| absolutenesses | wUabssolleeHeVsw |
| safEGuARDing | safEguARDiiTnm |
| FAded | NFAdekd |
| nanoGrams | nanoGrams |
| ABCDE | jMS |
| JAckeR | JAckerk |

Finally, the last modification function is shown in equation (41), which takes a random character of $w$ and places it at its end. This function is also executed for $|w| = 2$ times. The modification happens with $0.6 + R$ probability.

$$C_5(w) = \begin{cases} w \oplus atr\left(w, ran\left(0, |w| - 1\right)\right), & \text{if } ran(0,1) \geq 0.4 - R \\ w, & \text{otherwise} \end{cases} \tag{41}$$

The final modified string is given by $C_F$ in equation (42), where $N \times$ represents the number of times the associated function is called, where $N = |w|/2$. $N \times C_5$ means that $C_5$ is called $N$ times.

$$C_F(w) = N \times C_5\left(N \times C_4\left(C_3\left(N \times C_2\left(C_1(w)\right)\right)\right)\right) \tag{42}$$

After applying $C_F$ to string $w$, some instances such as the ones shown in the second column of Table 1 are produced. The entire process of modifying the generated string is also illustrated in Figure 6, given that each step depicted in the flowchart may or may not happen. Clearly, it is possible to change the degree of randomness of the generated dataset by adjusting the variable $R$.

**Figure 6** Order of the transformation on words

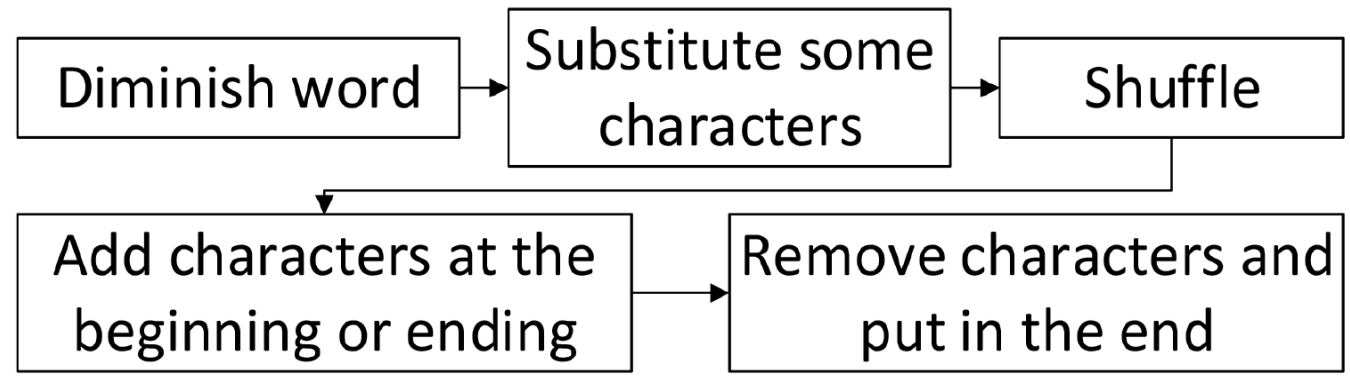

Finally, the strings of the dataset are then generated following the simple rule shown in equation (43) for string 1 ($w_1$) and equation (44) for string 2 ($w_2$). The variable $M$ stands for the maximum length of the string, $M = 14$ and $M = 200$ were used in the experiments.

$$W_1(M) = \bigoplus_{1}^{ran(1,M)} ran\left(\text{'}A\text{'}, \text{'}z\text{'}\right) \tag{43}$$

Assuming $w_1$ has been generated by function $W_1(M)$, such that $w_1 = W_1(M)$, then $w_2$ is returned by the function $W_2(w_1, M)$ based on $w_1$ when $w_2$ is supposed to be the same word as $w_1$. Otherwise, $w_2$ is randomly generated.

$$W_2\left(w_1, M\right) = \begin{cases} W_1,(M), & \text{if } ran(0,1) > 0.5 \\ C_F\left(w_1\right), & \text{otherwise} \end{cases} \tag{44}$$

For each experimental analysis $\{(M = 14, R = 0.5), (M = 14, R = 0.9), (M = 200, R = 0.0), (M = 200, R = 0.15)\}$ a dataset of 100,000 words was generated following the rules of the previous equations, from where the features proposed and discussed in this work are extracted. The first two experiments generate strings of up to 14 characters, extracts all the previously proposed features from each pair of words, and the performance of the features are analysed in groups, using several classification algorithms.

The second group of experiments ($M = 200$) considers strings of up to 200 characters. As the length of the strings is greater, the strings being compared are more random in relation to each other than in the 14 characters occasion. This is evident as equations $C_2$, $C_3$ and $C_5$ consider the length of the string as a parameter to perform more changes in the same. That mimics real situations. Longer texts that are supposed to be the same would probably suffer more changes than short versions of similar texts.

In these two major experiments, the $R$ variable shown in equations $C_{1...5}$, varies, in order to check whether features perform better or worse in different situations. In fact, features based on RLM can tolerate more randomness than features based on COM, as observed in the experiments. $R$ was altered with the intent of reinforcing this hypothesis. The higher the $R$ variable, the higher the randomness introduced while modifying $w_1$ to compose $w_2$. The degree of influence of $R$ differs regarding the two performed experiments ($M = 14$ and $M = 200$).

This synthetic dataset composes a binary classification problem. That is, one of the strings is always randomly generated. The second one is either

1 randomly generated

2 a modified version of the first.

Both occasions 1 and 2 occur with equal probability (50%). A binary label that represents whether the strings are supposed to be the same is associated to the comparison, as shown in Table 2. Labelled instances such as these are used to train the classification algorithms.

**Table 2** Instances for classification

| *String 1 ($w_1$)* | *String 2 ($w_2$)* | *Class or label* |
|---|---|---|
| kindheartedness | GindheaO | FALSE |
| naively | aiveyy | TRUE |
| earmark | JhluAmyyaCiTSm | FALSE |
| jREhjnABC | jREhnABC | TRUE |
| ideal | ijMS | FALSE |

### *4.1.1 Word comparison experiment*

Table 3 shows the accuracies of ten distinct classifiers when up to 14 characters long strings are compared. Accuracy is a rate given by equation (45),

$$Accuracy = \frac{TruePositives + TrueNegative}{Total\ \ population} \tag{45}$$

where true positives accounts for the cases where the classifier predicted that the strings being compared are the same and they actually are, while true negatives correspond to cases where the classifier predicted the strings are not the same and in fact they are not. The total population is equal to the total amount of comparisons or pairs of strings.

Used classifiers include random forest (Breiman, 2001), which is an ensemble of decision trees; Hoeffding tree (Hulten et al., 1990), which is also a decision tree algorithm; radial basis function (RBF) network (Poggio, 1990), which is a neural network; Naive Bayes (John and Langley, 1995), which is based on the Bayes theorem; decision table (Kohavi, 2005), which is a decision table algorithm; k-nearest neighbour (IBk in Weka) (Altman, 1992), which is a very simple and lazy algorithm that works with non-categorical variables and generates the predictive model during the classification phase; multilayer perceptron (Rosenblat, 1962), which is also a neural net; logistical model trees (LMT) (Landwehr et al., 2005), which is a combination of logistic regression and decision trees; and, finally, REPTree (Hall et al., 2009), which is a fast decision tree learner. The Weka (Hall et al., 2009; Rodrigues et al., 2017) framework was used to run the classification experiments.

**Table 3** Accuracies (%) for strings of up to 14 characters for $R = 0.5$

| _Algorithms_ | _Length_ | _LCS_ | _MCLCS_ | _MI_ | _Dist._ | _WMI_† | _COM_† | _RLM_† | _All_ | _Mean_• |
|---|---|---|---|---|---|---|---|---|---|---|
| LMT | 50.28 | 85.81* | 85.20* | 52.27* | 85.96* | 75.28* | _86.78_* | 82.45* | _87.53_* | _76.84_ |
| R. forest | 49.49* | 85.87* | 84.98* | 52.85* | 86.53* | 75.85* | 86.39* | 82.45* | 86.62* | 76.78 |
| REP tree | 50.06* | 85.82* | 85.19* | 52.73* | 85.75* | 74.90* | 86.56* | 82.50* | 87.12* | 76.74 |
| Multilayer P. | 50.12 | 85.47* | 84.71* | 51.49 | 85.06* | 74.06* | 86.42* | 82.35* | 87.15* | 76.31 |
| Ibk (k-NN) | 49.36* | 85.87* | 84.92* | 52.38* | 80.93* | 72.74* | 85.80* | 82.40* | 85.03* | 75.49 |
| Hoeff. tree | 49.76 | 85.56* | 84.48* | 51.38* | 78.93* | 66.39* | 84.84* | 81.98* | 86.03* | 74.37 |
| Dec. table | 50.32 | 84.83* | 84.55* | 50.43* | 72.67* | 62.48* | 86.27* | 82.51* | 85.34* | 73.27 |
| RBF net. | 50.29 | 83.51* | 79.41* | 50.35 | 75.10* | 55.88* | 81.18* | 78.89* | 85.42* | 71.11 |
| Naïve Bayes | 49.94 | 82.20* | 78.38* | 50.22 | 65.14* | 54.84* | 80.41* | 80.14* | 84.32* | 69.51 |
| ZeroR | 50.32 | 50.32 | 50.32 | 50.32 | 50.32 | 50.32 | 50.32 | 50.32 | 50.32 | 50.32 |
| Mean° | 49.96 | _84.99_ | 83.54 | 51.57 | 79.56 | 68.05 | 84.96 | 81.74 | _86.06_ | - |

Notes: †denotes groups of features proposed in this work.
*denotes accuracies that are statistically significant in comparison to ZeroR ($P$ value $\leq 0.01$).
°the 'mean' row corresponds to the average results of the classifiers (excluding ZeroR) regarding the group of features at the same column.
•the 'mean' column corresponds to the average of the results obtained by every group of features regarding the classifier at the same row.

All the experiments were performed using stratified ten-fold cross-validation and the parameters of the algorithms were tuned using the grid search algorithm available in Weka, which is a parameter optimiser. Grid search was set to run for approximately one hour in a reduced dataset for each classification algorithm and different $M$ and $R$ parameters. It is important to highlight that as the dataset is huge (containing 100,000 instances), 1% of difference in accuracy rates is equivalent to 1,000 instances being correctly classified.

**Table 4** List of features in each group

| *Feature group* | *Extracted features and used parameters* |
|---|---|
| Length | $\|w_1\|$; $\|w_2\|$; $\|w_2\| - \|w_1\|$; $abs(\|w_2\| - \|w_1\|)$ |
| LCS | $NLCS(w_1, w_2)$ |
| MCLCS | $NMCLCS_{0,0}(w_1, w_2)$, $NMCLCS_{0,1}(w_1, w_2)$; $NMCLCS_{\frac{\|w_1\|}{2},\frac{\|w_2\|}{2}}(w_1, w_2)$; $NMCLCS_{\forall n_1,\forall n_2}(w_1, w_2)$ |
| MI | $MI_0(w_1, w_2)$; $MI_1(w_1, w_2)$; $M_{I4}(w_1, w_2)$; $MI_{\forall d}(w_1, w_2)$ |
| Distance | $ModHam(w_1, w_2)$; $Lev_{w_1,w_2}$; $Dam_{w_1,w_2}$; $Dice(w_1, w_2)$ |
| WMI | $PWMI_0(w_1, w_2)$; $PWMI_1(w_1, w_2)$; $PWMI_4(w_1, w_2)$; $PWMIS(w_1, w_2)$ |
| COM | $COM\left(w_1, w_2, 0, \frac{\|w_1\|}{2}\right)$; $COP(w_1, w_2; 0)$; $COP(w_1, w_2, 1)$; $TPS(w_1, w_2)$; $COD(w_1, w_2)$ |
| RLM | $SO(w_1, w_2)$; $WSO(w_1, w_2)$; $MO(w_1, w_2)$; $MOML(w_1, w_2)$; $MORL(w_1, w_2)$; $MLMO(w_1, w_2)$; $RLMMCLCS(w_1, w_2)$ |

Notes: 1 Every group has its listed features plus the features of the 'length' group.
2 $MI_d$ means that $w_2$ is shifted using the SHIFT function with distance $d$ before computing MI.
3 Dice is not a distance, although it is in the 'distance' group.

ZeroR classifier was not mentioned before and deserves special attention. This classifier is the simplest classifier in Weka. It selects the most frequent label (or the mode) in the training dataset and use it to classify all unlabelled instances. If we have a binary classification problem where 51% of the training labels are equal to true, which implies that the remaining 49% are equal to false, ZeroR would classify any unlabelled instance as true, since true is the mode of the training dataset (51%). It deserves special attention because most significance tests performed in this work are done in relation to the ZeroR classifier, which is our null hypothesis. If no statistical significance is found in relation to ZeroR, that would be equivalent to saying that there is no statistical relationship between the analysed features and the label.

As it can be observed in Table 3, the classifier that obtained the highest averaged accuracy was LMT. LCS was the best group of features in average (84.99%). Still, features based on COM provided the best individual result (86.78%), rightly followed by the distance group (86.53%). In this case, the difference regarding these results may not be statistically significant. However, the features based on COM seem competitive enough. Reducing the number of features of the COM group, and thus increasing generalisation, could also improve its performance in this case.

Decision tree algorithms like C4.5, for instance, seek to make optimal splits in attribute values. Attributes that are more correlated with the prediction are split on first. Deeper in the tree, less relevant and irrelevant attributes are used to create predictions that may only be beneficial by chance in the training dataset, or sampling biased. This overfitting of the training data can negatively affect the modelling power of the method, decreasing the accuracy. However, since the approaches here are general, i.e., they readily work for different datasets, there is not much that can be done in this regard.

Table 4 shows the individual features for each group of features shown in Table 3. Every group contains the features described in this table along with the four features of the 'length' group. LCS group has five features:

1 $NLCS(w_1, w_2)$

2 $|w_1|$

3 $|w_2|$

4 $|w_2| - |w_1|$

5 $abs(|w_2| - |w_1|)$, where | | still represents the length of the string and abs function gets the absolute value.

Column 'all' in Table 3 stands for all features of all groups together. The column 'length' contains the described features associated to the length of the words (the four length features described in the previous paragraph).

**Table 5** Accuracies (%) for strings of up to 14 characters for $R = 0.9$

| *Algorithms* | *Length* | *LCS* | *MCLCS* | *MI* | *Dist.* | *WMI*† | *COM*† | *RLM*† | *All* | *Mean*• |
|---|---|---|---|---|---|---|---|---|---|---|
| LMT | 50.16 | 71.94⋆ | 71.82⋆ | 50.27 | 72.01⋆ | 65.04⋆ | *73.26*⋆ | 68.59⋆ | *73.74*⋆ | *66.31* |
| REP tree | 50.00 | 72.05⋆ | 71.52⋆ | 50.29 | 71.82⋆ | 64.91⋆ | 73.06⋆ | 68.60⋆ | 73.53⋆ | 66.20 |
| R. forest | 49.74⋆ | 71.99⋆ | 71.52⋆ | 50.36 | 71.87⋆ | 65.38⋆ | 72.80⋆ | 68.37⋆ | 72.96⋆ | 66.12 |
| Ibk (k-NN) | 49.73⋆ | 71.92⋆ | 71.46⋆ | 50.11 | 68.28⋆ | 64.18⋆ | 72.29⋆ | 68.32⋆ | 71.78⋆ | 65.34 |
| Multilayer P. | 50.00⋆ | 71.03⋆ | 70.90⋆ | 50.42 | 70.74⋆ | 62.60⋆ | 72.42⋆ | 67.63⋆ | 71.13⋆ | 65.22 |
| Dec. table | 50.14 | 71.68⋆ | 71.28⋆ | 50.27 | 62.83⋆ | 63.08⋆ | 72.80⋆ | 68.44⋆ | 73.07⋆ | 64.85 |
| Hoeff. tree | 50.17 | 71.82⋆ | 71.18⋆ | 50.34 | 66.08⋆ | 60.30⋆ | 72.42⋆ | 68.20⋆ | 72.76⋆ | 64.78 |
| RBF net. | 50.16 | 68.71⋆ | 69.12⋆ | 50.06 | 59.71⋆ | 52.63⋆ | 67.66⋆ | 66.72⋆ | 71.49⋆ | 61.81 |
| Naïve Bayes | 50.14 | 66.54⋆ | 66.66⋆ | 50.04 | 51.51⋆ | 51.33⋆ | 67.84⋆ | 66.68⋆ | 69.72⋆ | 60.05 |
| ZeroR | 50.20 | 50.20 | 50.20 | 50.20 | 50.20 | 50.20 | 50.20 | 50.20 | 50.20 | 50.20 |
| Mean° | 50.02 | 70.85 | 70.61 | 50.24 | 66.09 | 61.05 | *71.62* | 67.95 | *72.24* | - |

Notes: †denotes groups of features proposed in this work.
⋆denotes accuracies that are statistically significant in comparison to ZeroR ($P$ value $\leq 0.01$).
°the 'mean' row corresponds to the average results of the classifiers (excluding ZeroR) regarding the group of features at the same column.
•the 'mean' column corresponds to the average of the results obtained by every group of features regarding the classifier at the same row.

Table 5 shows the results for strings up to 14 characters still but regarding $R = 0.9$, which increments the randomness of the second string in relation to the first. Once again, LMT classifier outperformed the remaining. The best accuracy was also produced with features derived from the COM group. In this case, as opposed to the previous, COM based features produced the overall best accuracies (71.66%). This fact may indicate that COM based features perform better than LCS when more randomness between the words is introduced.

In Table 5, the COM group of features achieved the best accuracy per group: 73.26%. The second best accuracy per group was obtained with LCS: 72.05%. The difference between both accounts for more than 1000 instances that were correctly classified using the COM group instead of LCS. The probability of correctly classifying 1,000 instances by chance is $\left(\frac{1}{2}\right)^{1,000} \cong 9.3\times10^{-302}$, which is very unlikely. Thus, it is reasonable to argue that this difference is statistically significant. The use of several classifiers also reduces the sampling bias. When applying ten-fold cross validation, the instances are also randomly partitioned, which also mitigates the sampling problem.

In addition, in both experiments (Tables 3 and 5), MI based features could not provide valuable information to the classifiers. The improved MI, called WMI, in the other hand, improved the accuracy of the classifiers in comparison. However, both (WI and WMI) remain weaker than the remaining groups of features (except for the 'length' group). It is possible to say that the results obtained with every group except for 'length' and 'MI' are statistically significant.

### *4.1.2 Sentence comparison experiment*

For the second subgroup of experiments, strings of up to 200 characters were generated. In this case the strings represent sentences, as opposed to the previous set of experiments, where they represented words (they had up to 14 characters). Thus, the strings of this experiment contain a higher degree of randomness. Table 6 shows the obtained accuracies for $R = 0$. $R$ was set to zero to reduce the randomness, and that is reflected on the accuracies obtained, which were high.

In Table 6, the best averaged (97.81%) and individual (99.62%) accuracies were produced with the distance group, rightly followed by RLM. Although this experiment is not highly random ($R = 0$), the length of the strings increase the performance of the RLM group of features, since they are based on run lengths. If the strings are longer, a greater amount of long run lengths are present and that consequently increases the performance of this group. COM, LCS and MCLCS also achieved a good performance regarding this experiment, these three groups seem to work better when less randomness is introduced. The best classifier was again LMT but Random Forest can be considered as good as LMT in this case.

In the following experiment, the randomness is increased a little bit ($R = 0.15$). The obtained results are shown in Table 7. It can be stated that just by slightly increasing the $R$ variable, all accuracies drop substantially. That is due to the fact that the strings are longer (up to 200 characters) in relation to the previous set of experiments, and the length of the strings also influences on the randomness introduced during their generation, as previously addressed.

In this case, the mean result obtained with the RLM group substantially outperforms others. The highest accuracy was also obtained with the RLM group. Furthermore, LMT was once more the classification algorithm that outperformed the remaining.

**Table 6** Accuracies (%) for strings of up to 200 characters for $R = 0$

| *Algorithms* | *Length* | *LCS* | *MCLCS* | *MI* | *Dist.* | *WMI*† | *COM*† | *RLM*† | *All* | *Mean*• |
|---|---|---|---|---|---|---|---|---|---|---|
| LMT | 50.05 | 83.51★ | 98.55★ | 64.58★ | *99.62*★ | 75.18★ | 94.75★ | 99.61★ | *99.83*★ | *89.45* |
| R. forest | 50.27 | 81.82★ | 98.49★ | 65.05★ | 99.43★ | 76.48★ | 94.57★ | 99.55★ | 99.83★ | 89.40 |
| REP tree | 49.75 | 82.68★ | 98.35★ | 62.49★ | 99.17★ | 73.42★ | 94.34★ | 99.43★ | 99.73★ | 88.70 |
| Multilayer P. | 50.00 | 82.15★ | 97.95★ | 62.35★ | 99.11★ | 73.97★ | 94.52★ | 99.37★ | 99.52★ | 88.62 |
| Hoeff. tree | 49.78 | 82.29★ | 97.95★ | 59.80★ | 97.98★ | 69.13★ | 93.20★ | 98.72★ | 99.01★ | 87.26 |
| Ibk (k-NN) | 50.44? | 79.36★ | 90.98★ | 58.75★ | 98.83★ | 70.25★ | 92.06★ | 99.15★ | 99.37★ | 86.09 |
| Dec. table | 50.00 | 73.97★ | 95.18★ | 55.70★ | 97.96★ | 66.31★ | 92.73★ | 99.48★ | 99.03★ | 85.05 |
| Naïve Bayes | 49.98 | 71.07★ | 85.73★ | 53.99★ | 97.58★ | 55.94★ | 88.13★ | 89.36★ | 99.53★ | 80.17 |
| RBF net. | 50.06 | 72.93★ | 84.50★ | 54.70★ | 90.62★ | 56.59★ | 87.43★ | 80.85★ | 93.15★ | 77.60 |
| ZeroR | 50.00 | 50.00 | 50.00 | 50.00 | 50.00 | 50.00 | 50.00 | 50.00 | 50.00 | 50.00 |
| Mean° | 50.02 | 78.86 | 94.19 | 59.71 | *97.81* | 68.59 | 92.41 | 96.17 | *98.78* | - |

Notes: †denotes groups of features proposed in this work.
★denotes accuracies that are statistically significant in comparison to ZeroR (P value ≤ 0.01).
°the 'mean' row corresponds to the average results of the classifiers (excluding ZeroR) regarding the group of features at the same column.
•the 'mean' column corresponds to the average of the results obtained by every group of features regarding the classifier at the same row.

**Table 7** Accuracies (%) for words of up to 200 characters for $R = 0.15$

| *Algorithms* | *Length* | *LCS* | *MCLCS* | *MI* | *Dist.* | *WMI*† | *COM*† | *RLM*† | *All* | *Mean*• |
|---|---|---|---|---|---|---|---|---|---|---|
| LMT | 50.14 | 62.22★ | 67.60★ | 52.10★ | 67.10★ | 56.05★ | 64.85★ | *68.86*★ | *69.52*★ | *62.05* |
| REP tree | 49.87 | 62.00★ | 67.26★ | 51.70★ | 65.81★ | 55.74★ | 64.13★ | 68.26★ | 69.28★ | 61.56 |
| R. forest | 49.87 | 61.20★ | 66.03★ | 52.14★ | 65.73★ | 57.15★ | 63.50★ | 67.28★ | 68.70★ | 61.29 |
| Multilayer P. | 50.00 | 61.73★ | 65.47★ | 51.19★ | 65.70★ | 53.97★ | 64.00★ | 67.54★ | 67.02★ | 60.74 |
| Hoeff. tree | 49.86 | 61.86★ | 66.58★ | 51.39★ | 63.65★ | 53.59★ | 62.81★ | 67.88★ | 67.96★ | 60.62 |
| Dec. table | 50.16 | 60.07★ | 64.88★ | 50.84★ | 63.94★ | 53.47★ | 63.71★ | 67.97★ | 67.23★ | 60.25 |
| Ibk (k-NN) | 49.70 | 60.76★ | 62.34★ | 51.27★ | 64.47★ | 54.70★ | 62.34★ | 66.22★ | 65.72★ | 59.72 |
| Naive Bayes | 50.23 | 60.13★ | 61.34★ | 50.60 | 63.71★ | 50.57★ | 60.58★ | 62.12★ | 67.37★ | 58.52 |
| RBF net. | 50.10 | 59.93★ | 61.25★ | 50.49 | 60.34★ | 51.70★ | 60.82★ | 64.55★ | 65.63★ | 58.31 |
| ZeroR | 50.16 | 50.16 | 50.16 | 50.16 | 50.16 | 50.16 | 50.16 | 50.16 | 50.16 | 50.16 |
| Mean° | 49.99 | 61.10 | 64.75 | 51.30 | 64.49 | 54.10 | 62.97 | *66.74* | *67.60* | - |

Notes: †denotes groups of features proposed in this work.
★denotes accuracies that are statistically significant in comparison to ZeroR (P value ≤ 0.01).
°the 'mean' row corresponds to the average results of the classifiers (excluding ZeroR) regarding the group of features at the same column.
•the 'mean' column corresponds to the average of the results obtained by every group of features regarding the classifier at the same row.

In 3 out of 4 cases (Tables 3, 5 and 7), the robustness of proposed feature groups (RLM and COM) was statistically significant in comparison to the distance group using the LMT classifier. In Tables 3 and 5, P value was <0.001 and <0.001 for the COM group in contrast to the distance group, respectively. In Table 6, the RLM group tied with the distance group in terms of statistical significance. At last, in Table 7, the P value was <0.00000001 for the RLM group in contrast to the distance group.

### 4.1.3 Classification time comparison

Processing times required by classifiers were also analysed. Tables 8 and 9 contain the averaged times in seconds for training the predictive model and classifying the instances of one of the previous datasets, respectively. It is important to note however that these run times were provided by Weka (Java), which implies that they are not optimised. Fully optimised Java code is in general two times slower than optimised C code.

**Table 8** Mean time (s) for generating the predictive model over 100,000 instances

| *Algorithms* | *Length* | *LCS* | *MCLCS* | *MI* | *Dist.* | *WMI*† | *COM*† | *RLM*† | *All* | *Mean*• |
|---|---|---|---|---|---|---|---|---|---|---|
| Ibk (k-NN) | 0.03 | 0.03 | 0.04 | 0.04 | 0.04 | 0.04 | 0.04 | 0.03 | 0.05 | *0.04* |
| Hoeff. tree | 0.54 | 2.96 | 1.14 | 1.33 | 1.45 | 1.26 | 3.08 | 1.21 | 5.08 | 2.01 |
| RBF net. | 1.53 | 2.60 | 3.90 | 3.36 | 4.25 | 3.23 | 5.94 | 5.73 | 5.65 | 4.02 |
| Naive Bayes | 0.12 | 0.14 | 0.28 | 0.30 | 1.08 | 0.33 | 0.23 | 0.26 | 64.86 | 7.51 |
| REP tree | 1.86 | 3.30 | 5.47 | 9.61 | 4.62 | 5.16 | 9.18 | 14.31 | 21.00 | 8.28 |
| Dec. table | 1.12 | 1.70 | 4.12 | 14.52 | 4.84 | 13.82 | 4.10 | 5.21 | 62.29 | 12.41 |
| R. forest | 71.41 | 99.57 | 133.90 | 168.51 | 151.31 | 114.52 | 286.32 | 335.20 | 234.93 | 177.30 |
| Multilayer P. | 37.35 | 43.09 | 75.86 | 205.10 | 101.55 | 126.27 | 80.02 | 83.09 | 1,009.98 | 195.81 |
| LMT | 13.21 | 78.34 | 138.03 | 181.34 | 111.68 | 92.17 | 281.54 | 331.32 | 722.79 | 216.71 |
| Mean° | 14.13 | 25.75 | 40.30 | 64.90 | 42.31 | 39.64 | 74.49 | 86.26 | 236.29 | - |

Notes: †denotes the groups of features proposed in this work.
°the 'mean' row corresponds to the average results of the classifiers (excluding ZeroR) regarding the group of features at the same column.
•the 'mean' column corresponds to the average of the results obtained by every group of features regarding the classifier at the same row.

The results regarding k-NN or Ibk deserve special attention. k-NN is a lazy algorithm and due to that fact it 'generates its predictive model during the classification phase'. Therefore, the algorithm 'generates the predictive model' over and over again for each instance, undermining the overall classification time. This is also the reason why k-NN was the fastest regarding the predictive model generation, but it is due to the fact that it does not generate any model during that phase. LMT is significantly slower than other algorithms to generate the predictive model. It may be more valuable to use a faster classifier in detriment of the accuracy if a faster classification is desired. This trade-off should be analysed for each specific case.

It is also evident that when all the features are together ('all' column), all the algorithms significantly require more time to generate the predictive model (exceptions are k-NN and RBF network, which did not change substantially). However, the classification time does not vary as much for some algorithms, independently of the amount of features. Nevertheless, extracting more features from the strings require more time. Thus, this trade-off should also be analysed carefully, and that is why we perform a time analysis experiment of each feature in the next subsection.

**Table 9** Mean time (s) for classifying other 100,000 instances

| *Algorithms* | *Length* | *LCS* | *MCLCS* | *MI* | *Dist.* | *WMI*† | *COM*† | *RLM*† | *All* | *Mean*• |
|---|---|---|---|---|---|---|---|---|---|---|
| LMT | 0.36 | 0.35 | 0.35 | 0.88 | 0.17 | 0.67 | 0.46 | 0.46 | 0.81 | *0.50* |
| Dec. table | 0.52 | 0.50 | 0.45 | 0.80 | 0.17 | 0.85 | 0.43 | 0.48 | 0.91 | 0.57 |
| REP tree | 0.37 | 0.44 | 0.56 | 0.80 | 0.40 | 0.76 | 0.55 | 0.52 | 0.75 | 0.57 |
| RBF net. | 0.61 | 0.86 | 1.07 | 1.23 | 0.97 | 1.30 | 1.05 | 0.99 | 1.58 | 1.07 |
| Hoeff. tree | 0.61 | 1.03 | 0.99 | 1.48 | 0.90 | 1.10 | 1.81 | 1.86 | 1.58 | 1.26 |
| Multilayer P. | 0.35 | 4.58 | 0.71 | 0.98 | 0.27 | 0.92 | 0.64 | 0.59 | 3.87 | 1.43 |
| Naive Bayes | 0.60 | 0.62 | 0.83 | 1.46 | 0.34 | 1.42 | 0.88 | 0.90 | 35.14 | 4.69 |
| R. forest | 12.16 | 13.49 | 20.95 | 14.59 | 19.30 | 15.78 | 23.44 | 19.80 | 17.43 | 17.44 |
| Ibk (k-NN) | 325.91 | 385.35 | 497.88 | 476.25 | 618.84 | 480.94 | 430.31 | 474.24 | 720.05 | 489.97 |
| Mean° | 37.94 | 45.25 | 58.20 | 55.39 | 71.26 | 55.97 | 51.06 | 55.54 | 86.90 | - |

Notes: †denotes the groups of features proposed in this work.
°the ‘mean’ row corresponds to the average results of the classifiers (excluding ZeroR) regarding the group of features at the same column.
•the ‘mean’ column corresponds to the average of the results obtained by every group of features regarding the classifier at the same row.

### *4.1.4 Feature extraction time comparison*

Table 10 shows the averaged times in nanoseconds for extracting each feature. The values in the table are the mean of 100,000 extractions over different instances of one of the previous datasets (200 characters and $R$ = 0.15). The features based on COM and RLM are derived from these constructed matrices, and these matrices are constructed once. As from them, several features can be extracted without the requirement of regenerating these matrices. Thus, the time to generate the matrices is accounted just once. Furthermore, the proposed features were implemented and run in Java and they are not optimised, therefore, the times can still be improved substantially.

### *4.1.5 Attribute ranking*

Finally, an analysis of the quality of the features was also performed using Weka’s classifier attribute evaluator, which evaluates the worth of an attribute using a specified classifier. LMT was selected as the classifier, since it was the best classifier in previous experiments. The evaluation measure was set to accuracy. A ranking of the attributes was generated, where the one that most positively influences accuracy is ranked in the beginning. Tables 11 and 12 show the rank of the 15 most valuable attributes for strings of length up to 14 and $R$ = 0.9 as well as for strings up to 200 characters and $R$ = 0.15, respectively.

**Table 10** Mean time (ns) required to extract each feature, averaged for 100,000 instances

| *Feature group* | *Feature: mean time (ns)* | *Group time (ns)** |
|---|---|---|
| Length | $\lvert w_1\rvert$: 370.40; $\lvert w_2\rvert$: 295.48; $\lvert w_2\rvert - \lvert w_1\rvert$: 317.91; $abs(\lvert w_2\rvert - \lvert w_1\rvert)$: 322.62 | 1,306.41 |
| LCS | $NLCS(w_1, w_2)$: 2,048.16 | 2,048.16 |
| MCLCS | $NMCLCS_{0,0}(w_1, w_2)$: 523.72; $NMCLCS_{0,1}(w_1, w_2)$: 432.34; $NMCLCS_{\frac{\lvert w_1\rvert}{2},\frac{\lvert w_2\rvert}{2}}(w_1, w_2)$: 477.72; $NMCLCS_{\forall n_1,\forall n_2}(w_1, w_2)$: 2552.20 | 3,985.98 |
| MI | $MI_0(w_1, w_2)$: 4,330.20; $MI_1(w_1, w_2)$: 2,923.17; $MI_4(w_1, w_2)$: 2,801.42; $MI_{\forall d}(w_1, w_2)$: 17,974.44 | 28,029.23 |
| Distance | $ModHam(w_1, w_2)$: 1,368.39; $Lev_{w_1,w_2}$: 984.33; $Dam_{w_1,w_2}$: 4,100.81; $Dice(w_1, w_2)$: 3,112.76 | 9,566.29 |
| WMI | $PWMI_0(w_1, w_2)$: 2,939.77; $PWMI_1(w_1, w_2)$: 2,812.76; $PWMI_4(w_1, w_2)$: 2,798.27; $PWMIS(w_1, w_2)$: 17,727.56 | 26,278.36 |
| COM† | $COM\left(w_1, w_2, 0, \frac{\lvert w_1\rvert}{0}\right)$: 709.14; $COP(w_1, w_2, 0)$: 782.15; $COP(w_1, w_2, 1)$: 644.40; $TPS(w_1, w_2)$: 1,791.27; $COD(w_1, w_2)$: 3,599.83 | 12,079.24 |
| RLM‡ | $SO(w_1, w_2)$: 500.08; $WSO(w_1, w_2)$: 512.16; $MO(w_1, w_2)$: 507.55, $MOML(w_1, w_2)$: 491.48; $MORL(w_1, w_2)$: 512.35; $MLMO(w_1, w_2)$: 521.02; $RLMMCLCS(w_1, w_2)$: 522.18 | 20,271.57 |

Notes: †the COM group requires an additional 4,552.45 ns (averaged across 100,000 instances) for building the COM.
‡the RLM group requires an additional 16,704.75 ns (averaged across 100,000 instances) for building the RLM.
*group time column already contains these additional times described above.

It can be seen in Tables 11 and 12 that features derived from COM and RLM were the two most valuable features. As expected, RLM seems to perform better than COM in more random data/longer strings. Moreover, MI and Length based features appeared at last in the ranking, being the least valuable.

## 4.2 Plagiarism dataset experiment

Besides the synthetic generation of data, we also experiment with the dataset (Clough and Stevenson, 2011) provided by Clough and Stevenson. The authors created a corpus that can be used in the development and evaluation of plagiarism detection systems that reflect the types of plagiarism performed by students in an academic setting. In what follows, the original dataset description provided by the authors is presented.

This is a direct quote from the work that originally proposed this dataset:

> "A set of five short answer questions on a variety of topics that might be included in the computer science curriculum were created by the authors. For each of these questions a set of answers were obtained using a variety of approaches, some of which simulate cases where the answer is plagiarised and others that simulate the opposite. To simulate plagiarism, the authors used a suitable Wikipedia entry as source text from which participants plagiarised. Four levels of plagiarism are represented in the corpus:
>
> 1 *Near copy:* "Participants were asked to answer the question by simply copying text from the relevant Wikipedia article (i.e., performing cut-and-paste actions). No instructions were given about which parts of the article to copy (selection had to be performed to produce a short answer of the required length, 200–300 words)."

2 *Light revision:* "Participants were asked to base their answer on a text found in the Wikipedia article and were, once again, given no instructions about which parts of the article to copy. They were instructed they could alter the text in some basic ways including substituting words and phrases with synonyms and altering the grammatical structure (i.e., paraphrasing). Participants were also instructed not to radically alter the order of information found in sentences."

3 *Heavy revision:* "Participants were once again asked to base their answer on the relevant Wikipedia article but were instructed to rephrase the text to generate an answer with the same meaning as the source text, but expressed using different words and structure. This could include splitting source sentences into one or more individual sentences, or combining more than one source sentence into a single sentence. No constraints were placed on how the text could be altered."

4 *Non-plagiarism:* "Participants were provided with learning materials in the form of either lecture notes or sections from textbooks that could be used to answer the relevant question. Participants were asked to read these materials and then attempt to answer the question using their own knowledge (including what they had learned from the materials provided). They were also told that they could look at other materials to answer the question but were explicitly instructed not to look at Wikipedia.""

**Table 11** Feature ranking (length 14 and $R = 0.9$)

| *Rank* | *Feature* | *Group* |
|---|---|---|
| 1 | $\frac{TPS(w_1, w_2)}{\lvert w_2 \rvert}$ | COM |
| 2 | $NLCS(w_1, w_2)$ | LCS |
| 3 | $NMCLCS_{\forall n_1, \forall n_2}(w_1, w_2)$ | MCLCS |
| 4 | $\frac{SO(w_1, w_2)}{\lvert w_2 \rvert}$ | RLM |
| 5 | $COD(w_1, w_2)$ | COM |
| 6 | $RLMMCLCS(w_1, w_2)$ | RLM |
| 7 | $MOML(w_1, w_2)$ | RLM |
| 8 | $MLMO(w_1, w_2)$ | RLM |
| 9 | $MORL(w_1, w_2)$ | RLM |
| 10 | $WSO(w_1, w_2)$ | RLM |
| 11 | $MO(w_1, w_2)$ | RLM |
| 12 | $PWMIS(w_1, w_2)$ | WMI |
| 13 | $NMCLCS_{0,0}(w_1, w_2)$ | MCLCS |
| 14 | $COP(w_1, w_2, 1)$ | COM |
| 15 | $COP(w_1, w_2, 0)$ | COM |

Notes: 1 Ranking 1 stands for the best or most valuable feature.
2 Features were evaluated using classifier attribute evaluator and LMT classifier.

**Table 12** Feature ranking (length 200 and $R = 0.15$)

| *Rank* | *Feature* | *Group* |
|---|---|---|
| 1 | $MORL(w_1, w_2)$ | RLM |
| 2 | $RLMMCLCS(w_1, w_2)$ | RLM |
| 3 | $NMCLCS_{\forall n_1, \forall n_2}(w_1, w_2)$ | MCLCS |
| 4 | $Dice(w_1, w_2)$ | Distance |
| 5 | $SO(w_1, w_2)$ | RLM |
| 6 | $\frac{TPS(w_1, w_2)}{\lvert w_2 \rvert}$ | RLM |
| 7 | $\frac{SO(w_1, w_2)}{\lvert w_2 \rvert}$ | RLM |
| 8 | $COP(w_1, w_2, 0)$ | COM |
| 9 | $COP(w_1, w_2, 1)$ | COM |
| 10 | $COM\left(w_1, w_2, 0, \frac{\lvert w_1 \rvert}{2}\right)$ | COM |
| 11 | $TPS(w_1, w_2)$ | COM |
| 12 | $WSO(w_1, w_2)$ | RLM |
| 13 | $MOML(w_1, w_2)$ | RLM |
| 14 | $MO(w_1, w_2)$ | RLM |
| 15 | $NMCLCS_{0,0}(w_1, w_2)$ | MCLCS |

Notes: 1 Ranking 1 stands for the best or most valuable feature.
2 Features were evaluated using classifier attribute evaluator and LMT classifier.

This dataset contains four possible labels:

1 near copy

2 light revision

3 heavy revision

4 non-plagiarism.

The same features that were extracted in the previous datasets were also extracted in this dataset. The employed classification algorithms were also the same. The only difference lays on the fact that there this is a multi-label problem instead of a binary classification problem. The stratified ten-fold cross validation is once again used.

**Table 13** Comparison of accuracies using the Wikipedia plagiarism dataset

| *Author* | *Accuracy* |
|---|---|
| Clough and Stevenson (2011) | 66.31 |
| Chong et al. (2014) | 70.52 |
| *This work* | *84.21* |

*Source:* Clough and Stevenson (2011)

This dataset was pre-processed, adhering to the proposed methodology. At first, all texts were scanned and every character that does not correspond to a letter or a blank space was removed from them. Later, every different word was associated to a different character (the UTF16 encoding was used). Each pre-processed text ($w_1$) was then associated to the original Wikipedia passage ($w_2$). The labels for each $w_1$ are also provided by the authors. Thus, the problem is to compare $w_1$ and $w_2$ and to infer which one of the four labels corresponds to the comparison of texts $w_1$ and $w_2$.

Clough and Stevenson (2011) reported in their original work an accuracy of 80% (which can be found in Table 5 of their work) for this dataset using a simple set of features. However, along with their reported accuracy, the authors also provide their confusion matrix (Table 6 of their work). The sum of the numbers in its diagonal line represents the number of correctly classified instances, and they sum to 63 occasions in 95, which corresponds to an accuracy of 66.31%.

Initially, we extracted the same features proposed by Clough and Stevenson (2011) and experimented them with some classifiers. However, we were not able to achieve their reported accuracy of 80%. The closest we could get to their report was 70%. This fact further suggested to look at their confusion matrix, which revealed the inconsistency. This probably indicates that their reported accuracy (80%) is incorrect, and that their method correctly labels just 66.31% of instances in their dataset.

Table 13 compares the accuracies obtained using the Wikipedia plagiarism dataset provided by Clough and Stevenson (2011). Chong et al. (2014), which are also listed in Table 13, use a combination of features such as trigrams, bigrams, LCS and dependency relations, also coupled with a classification algorithm (Naive Bayes).

In order to obtain an accuracy of 84.21%, we use the Vote classifier available in Weka. This classifier counts the votes provided by one or more classifiers for each instance, and uses this information in the labelling process. In this case, we use a combination of four classifiers and count their votes:

1 multilayer perception

2 simple logistic

3 Ibk – or kNN

4 J48 – or C4.5.

The combination rule for the votes was the product of the class probability provided by each classifier. This complex configuration was found by Auto-Weka and was better than our manual adjustments. However, it is also possible to achieve interesting results using a single classifier and not an ensemble. We were able to achieved up to 76.84% of accuracy using just the LWT classifier alone, which is already greater than the state-of-the-art.

Furthermore, a dimensionality reduction was also performed before proceeding with the classification phase, in order to speed up training phases and improve obtained accuracies. The result obtained with our methodology outperforms both works in the literature that experimented with this plagiarism dataset. This dimensionality reduction selected the rlmMCLCS and rlm-MORL features along with the dice distance.

**Table 14** Confusion matrix obtained with the proposed approach

| - | *Non-plagiarism* | *Light* | *Heavy* | *Cut* |
|---|---|---|---|---|
| Non-plagiarism | 38 | 0 | 0 | 0 |
| Light | 4 | 15 | 0 | 0 |
| Heavy | 4 | 0 | 15 | 0 |
| Cut | 7 | 0 | 0 | 12 |

**Table 15** Obtained indices per class

| - | *TP* | *Precision* | *ROC area* | *Recall* |
|---|---|---|---|---|
| Non-plagiarism | 1.0 | 0.717 | 0.933 | 1.0 |
| Light | 0.789 | 1.0 | 0.921 | 0.789 |
| Heavy | 0.789 | 1.0 | 0.997 | 0.789 |
| Cut | 0.632 | 1.0 | 0.940 | 0.632 |

**Table 16** Feature ranking (Wikipedia plagiarism dataset)

| *Rank* | *Feature* | *Group* |
|---|---|---|
| 1 | $MCLCS(w_1, w_2)$ | RLM |
| 2 | $MORL(w_1, w_2)$ | RLM |
| 3 | *Dice* | DIST |
| 4 | $NormalisedMCLCS(w_1, w_2)$ | RLM |
| 5 | $NMCLCS_{0,0}(w_1, w_2)$ | MCLCS |
| 6 | $NLCS(w_1, w_2)$ | LCS |
| 7 | $WSO(w_1, w_2)$ | RLM |
| 8 | $SO(w_1, w_2)$ | RLM |
| 9 | $COP(w_1, w_2, 1)$ | COM |
| 10 | $COP(w_1, w_2, 0)$ | COM |
| 11 | *Damerau* | DIST |
| 12 | $TPS(w_1, w_2, 0)$ | COM |
| 13 | *Levenshtein* | DIST |
| 14 | $COD(w_1, w_2)$ | COM |
| 15 | $COM(w_1, w_2, 0, 0)$ | COM |

Notes: 1 Ranking 1 stands for the best or most valuable feature.
2 Features were evaluated using classifier attribute evaluator and the vote classifier previously described in this section.

Table 14 illustrates the confusion matrix obtained in our experiments. Table 15, on the other hand, illustrates indices obtained per class.

At last, we show in Table 16 a ranking that was generated using the previously described Vote classifier. There is a predominance of RLM-related features in this problem among all evaluated features, proving the description power of this proposal.

## 5 Conclusions

This work proposes a framework for string similarity computation. The framework leans on the analysis of features extracted from two strings and its association to predefined labels that indicate the degree of similarity of these strings. All the features used in this work are built upon statistical or logical rule-based concepts. Therefore, the proposed framework is suitable for working with any language or type of text.

In terms of features, three novel groups are proposed. The first group is an extension of the mutual information measure where a weight factor is adopted. The second group of features stems from COMs, usually applied in image processing. At last, the third group contains features extracted from an adapted concept of the run length matrix. Both COMs and RLMs are widely used in visual computing, but have never been explored in string similarity computation.

The similarity problem was posed as a classification problem, where two strings are compared and a label determines whether the strings are the same. A vast extent of experiments was performed, where several degrees of randomness and string lengths were evaluated. In this first set of synthetic experiments, the proposed COM and RLM groups outperformed all groups of features in 3 out of 4 occasions. In the second set of experiments, where a real text plagiarism dataset is used, the proposed RLM group was more valuable than the remaining groups of features. Our methodology also obtained the best accuracy in the literature for this dataset.

It is possible to infer from the results reported in this work that mutual information may not be an efficient measure for string comparison. It is better to employ the proposed WMI instead. The weighted MI, however, is also not as good as the other groups. Features based on the length of the string are also not sufficient to provide trustworthy information in terms of classification.

In summary,

1 LCS, MCLCS, COM and distance groups contain features that are best suited for subtle randomness. The degree of randomness refers to the degree of difference between the strings being compared, two strings that are supposed to be the same but differ substantially would be substantially random between themselves. The distance group seems to tie with COM, where both outperform LCS and MCLCS in the experiments. MCLCS particularly works better than LCS for longer strings.

2 It is also possible to conclude that RLM, distance and MCLCS groups are best suited for more randomness. The RLM group was substantially better than these other two, proving the importance of this work proposal. RLM would probably outperform all the remaining groups in tasks such as text comparison, entailment or anything that involves longer strings or higher degrees of randomness.

Regarding the performance of classification algorithms, LMT was the one that obtained the best accuracies in our synthetic datasets, rightly followed by Random Forest. However, these algorithms are slow in the generation of the predictive model and they should be avoided if the application requires regenerating the predictive model over and over again. If the case is exactly the opposite, assuming that the predictive model is generated once, it is then recommended to use LMT. Decision table and Naive Bayes were the best classifiers regarding the plagiarism dataset. In the plagiarism dataset, an ensemble classifier that combined the votes (product of the probabilities) of 4 other classifiers was used.

As a final remark, this work did not explore the full potential of COM and RLM in string similarity computation. Several other features based on these matrices can still be proposed and further explored as future work. Finally, the datasets and source code used in this work can be found in Rodrigues (2020).